# Application of Enhanced-2D-CWT in Topographic Images for Mapping Landslide Risk Areas

Victor Vermehren Valenzuela[1,2], Rafael Dueire Lins[2], Hélio Magalhães de Oliveira[2]

[1]Universidade Estadual do Amazonas, Manaus-AM, Brazil
[2]Universidade Federal de Pernambuco, Recife-PE, Brazil
*vvv@netium.com.br, rdl@ufpe.br, hmo@ufpe.br*

**Abstract** – There has been lately a number of catastrophic events of landslides and mudslides in the mountainous region of Rio de Janeiro, Brazil. Those were caused by intense rain in localities where there was unplanned occupation of slopes of hills and mountains. It became thus imperative creating an inventory of landslide risk areas in densely populated cities. This work presents a way of demarcating risk areas by using the bidimensional Continuous Wavelet Transform (2D-CWT) applied to high resolution topographic images of the mountainous region of Rio de Janeiro.

## 1. Introduction

Landslide is a phenomenon caused by the slipping of solid materials such as soils, rocks, vegetation, and construction debris along the terrain slopes, called hillsides. In general, it occurs in mountainous areas, where the original vegetation was removed. The landslide differs from erosion by the amount of mass transported at great speed. Such a phenomenon in populated areas may cause tragedies yielding losses both of lives as economic.

Unplanned urban occupation in cities of developing countries has a recurrent occurrence, overall in poor areas that become very densely inhabited. Cities in expansion move onto topographically inclined and geologically unstable land, such as the slopes of hills. Brazil, predominantly situated in the tropics, often has summer storms, which sometimes reaches very high rainfall indexes. During such a period, hillsides have the potential risk of landslide. In Brazil, especially in the state of Rio de Janeiro, tragedies caused by landslides on hills and mountainous urban areas have been occurring with alarming frequency in recent years. Unfortunately, they happened with high proportions in the first week of 2011, considered the worst natural accident in Brazil, and in a smaller scale again in the second week of 2013. To worsen such a tragic scenario, these accidents often happen in overcrowded areas inhabited by the poorest population of the city. Thus, the creation of a detailed inventory mapping areas at risk of landslides is the first step towards implementation of public policies for the urban occupation [1]. Traditional methods of mapping hillsides include digitization of topographic maps for regular and irregular contours, aerial photo interpretation, and the direct observation in the field of the morphology of the slopes [2]. However, each of these methods has limitations that can affect the accuracy of the resulting map such as resolution, vegetation, rugged terrain, different map drawing criteria, and lack of spatial and temporal patterns [3].

Currently the demand for models of high-resolution ground mapping produced from airborne measurements has increased. Although laser is not a new technology, its use in the acquisition of spatial data is somewhat recent. Its employ in systems such as LiDAR (Light Detection And Ranging) has showed a fine ability to acquire a large amount of information in a short time. It is a remote sensing technology with the aim of measuring the distance to a target or object of interest, using laser pulses. It can be used in collecting data for terrestrial digital modeling (DTM) and digital elevation models (DEM) where traditional methods were not adequate, especially in areas of difficult access [4].

Wavelets have lately gained prolific applications throughout an amazing number of areas, especially in Physics and Engineering. Both Continuous and Discrete Wavelet transforms (CWT and DWT, respectively) have emerged as a definitive tool of signal processing analysis and have proven to be superior to classical Fourier analysis in many situations. Wavelet analysis for their ability to examine the signal simultaneously in time and frequency has generated a number of sophisticated methods based on wavelets, and among them those who study the surface morphology of the seas [5] and land [6]. In the latter, despite the normalized power of terrains, the used wavelet made no energy normalization to increase the accuracy in the results. This paper applies the continuous wavelet transform (CWT) with the normalized wavelet to extract the topographical features of slopes from high resolution topographic data of through the power spectral variations within specific spatial frequency ranges and compared with the Fourier transform [7-8].

Due to the high latent risk of accidents, the selected area is the mountainous region of Rio de Janeiro - which have available inventory of high-resolution maps DEM obtained by LiDAR in [9]. There are also sites independently detailing the region studied. Figure 1 shows the map of the study areas.

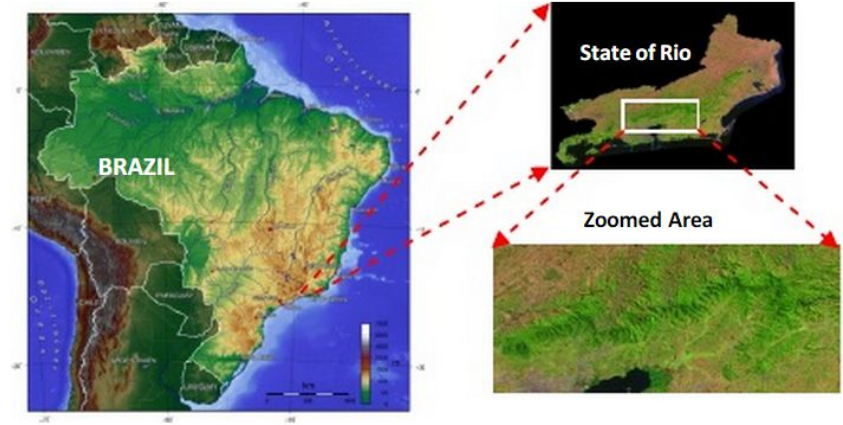

**Fig. 1.** Map of high resolution DEM of the Mountain Region – Rio de Janeiro obtained with LiDAR, INPE Database, area 22S435.

## 2. The algorithm HillShade and Bidimensional Continuous Walelet Transform

The application of digital image processing in this study is divided into two parts. First, one makes the application of continuous wavelet transform (2D-CWT) to quantify the topographic expressions of deep slopes and geologically unstable areas, mapping their locations in the areas of interest. The 2D-CWT transforms data into spatial frequency-position space, providing information about how the amplitude is distributed on spatial frequency at each position in the image. As the 2D DFT transformed the space domain into the frequency domain, it also provides consolidated information on how the amplitude of the topographic surface is distributed over a frequency spectrum that is used as a reference for comparison of results [10]. The other technique is the shade of hillsides [11], whose goal is to improve the display resolution greatly increasing the perception of a surface for analysis or graphical display. The foundation of this process is to apply the non-weighted gradient between neighboring image points (in this case, a topographic mapping). The result is a collection of vectors pointing in the direction of increasing values of the function $F$ (altimetric values). The equation gradient used in in-line functions (from MatLab$^{TM}$ [12], for example) is

$$\nabla F = \frac{\delta F}{\delta x}\hat{i} + \frac{\delta F}{\delta y}\hat{j} \quad (1)$$

This information is applied to a function that gets the hypothetical illumination surface by determining the illumination values for each cell of a matrix. This is done by creating a position for a hypothetical light source and calculating the illumination values for each cell in relation to its neighboring cells following the azimuth and slope parameters chosen. The function of shading (h) is given below.

$$h = 255.((\cos(zenith).\cos(slope)) + \sin(zenith).\sin(slope).\cos(azimute))) \quad (2)$$

The high resolution DEM map of the region of study with 235º azimuth and elevation 45º is shown in Figure 2.

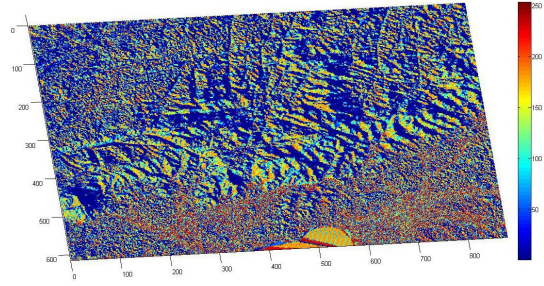

**Fig. 2**. Mountain region of Rio de Janeiro (referred in Fig.1) with application of shading algorithm.

The equation of the 2D CWT of a continuous signal, $g(x,y)$ with respect to wavelet function is defined as [13]

$$CWT(a,b,s) = \frac{1}{s} \int_{-\infty}^{\infty} \int_{-\infty}^{\infty} g(x,y)\psi_{abs}^{*}(x,y)dxdy, \qquad (3)$$

where $\psi$ represents the family of wavelets with scale parameter $s$ and location $(a, b)$. The 2D-CWT is the convolution of $g$ and $\psi$, and thus the transform coefficients CWT $(a, b, s)$ provide a measure of the degree of correlation between wavelet $\psi$ and $g$ data at each point. When the value of $s$ is high, $\psi$ is dispersed and takes into account the characteristics of wavelength $g$; when the value of $s$ is low, $\psi$ is best localized in space and achieves a better correlation with contours $g$.

The 2D-CWT can also be estimated within a power spectrum calculating its variance on points $Na \times Nb$ in each scale of the energy wavelet function [13]:

$$\sigma^2_{CWT}(s) = \frac{1}{2N_a N_b} \int\int |CWT(a,b,s)|^2 dadb \qquad (4)$$

Equation (4), similarly to the Fourier power spectrum, provides a general measure of how the amplitude topographical features changes with frequency throughout the data set. However, as in Figure 3, a wavelet analysis of a given scale typically a bandpass of frequencies cantered on a main frequency. Moreover, the DFT transform analyzes each frequency separately. Thus, this has the effect of smoothing the peaks in the power spectrum wavelet compared to its Fourier similar [14].

In the analysis, as its shape resembles the contours of the topography, the 2D Mexican Hat wavelet was used as wavelet mother, whose analytic expression is given by the following equations (time and frequency domains). Note that to ensure that wavelet transforms (3) at each scale $s$ are directly comparable with each other and to other series of transformations, the 2D Mexican Hat wavelet was normalized to have unitary energy:

$$\psi(x,y) = \frac{1}{2.507}(2 - x^2 - y^2) e^{-\frac{1}{2}(x^2+y^2)} \quad \text{and} \quad \Psi(u,v) = \frac{2}{\sqrt{3}\sqrt[4]{\pi}} \sqrt{32} \pi^{\frac{5}{2}} (u^2 + v^2) e^{-2\pi(u^2+v^2)} \qquad (5)$$

Figure 3 shows the time and frequency domain plots of the normalized 2D Mexican hat wavelet ($E=\sigma^2=s=1$).

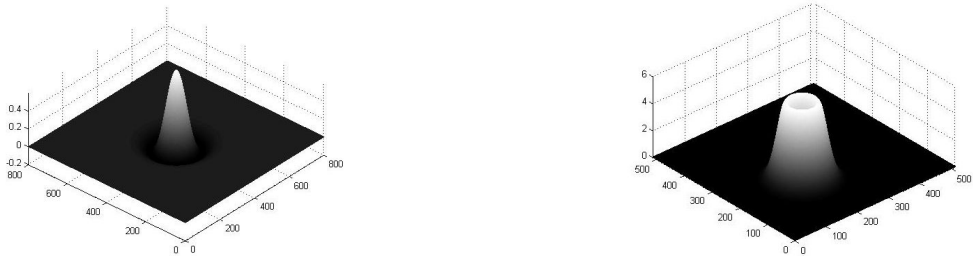

Fig. 3. (*left*) 2D Mexican wavelet, (*right*) its Fourier transform.

The Mexican hat is derived by the second derivative of a Gaussian function and has a wavelength $2\pi s/(5/2)^{1/2}$ times the grid spacing [15]. Performing the convolution of wavelet Mexican 2D with the topographic data one obtains the derived amplitude values and specific wavelengths, including characteristic features of hillsides.

The Fourier analysis of two-dimensional topographic images provides quantitative information on the amplitude, orientation, and shape of periodic information on a range of spatial frequencies [10]. The DFT of a 2D data set $f(x,y)$ with $M \times N$ nodes uniformly spaced and their respective power spectrum estimated by the DFT periodogram [16]

$$F(k,l) = \sum_{m=0}^{M-1}\sum_{n=0}^{N-1} f(m,n) e^{-j2\pi\left(\frac{k}{M}m + \frac{l}{N}n\right)} \quad \text{and} \quad p_{DFT} = \frac{1}{M^2 N^2} |F(k,l)|^2 \tag{6}$$

Whereas mapping landslides the periodogram provides a measure of amplitude of topographical features within a DEM in orientations and specific wavelength.

## 3. Results

In order to determine the characteristic wavelength of deep slopes of the studied area, first small areas representing rugged terrains and land smoothly (plans) were selected. Rio de Janeiro and the Northwest portion of Petrópolis, a city with clear mountains and hills were used as examples of hilly areas (Fig. 4a). The area west of Rio de Janeiro (Fig. 4b) was used as a representative of the flat seaside region.

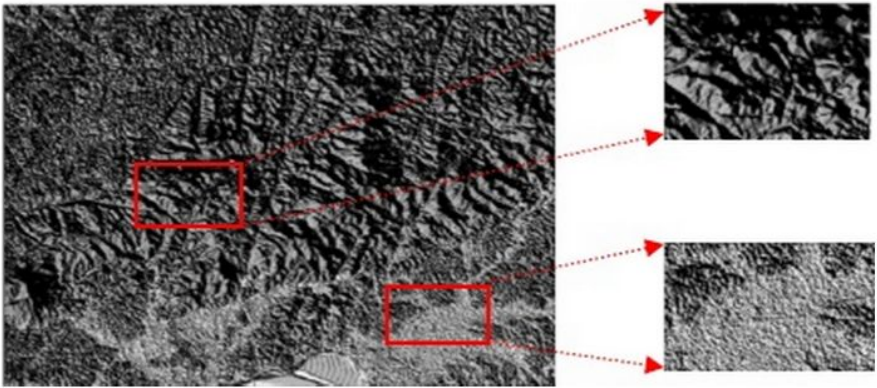

**Fig 4.** Left: LiDAR DEM-Map + shading.
(Right-top) portions of mountain slopes and (Right-bottom) relatively flat terrain.

Figure 5 presents the generated power spectra CWT and reference Fourier each land area by using Equations (4) and (6). The next procedure was to normalize the spectrum of these slopes by its corresponding spectrum in low (flat) lands in order to highlight more clearly the frequency bands over which deep slopes tend to concentrate power spectra. The two normalized spectra produce **well-defined peaks,** which indicate the wavelength characteristic of slopes at risk of landslides. The peaks of the spectrum indicate a wavelength of ~ 2222 m for the mountainous area of Rio de Janeiro, with negligible difference in position of the peaks between the CWT and Fourier spectra (Fig. 5).

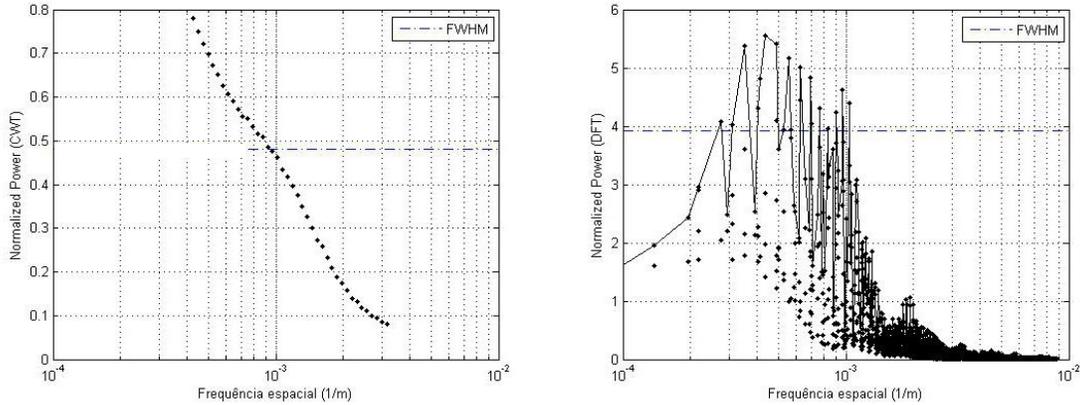

**Fig. 5**. Spectral Energy indicating characteristics frequency bands of the study area.
(left) the spectral power CWT and (right) the Fourier power spectrum.

As may be seen in Figure 5, the traces of the slopes in the studied area are within a range of wavelengths, thus the value of the width at half of the maximum (FWHM) of each peak was taken as a single measure of the spreading of the spectrum. The characteristic band of spatial frequencies obtained was used in the charts. In the mountainous region of Rio de Janeiro, the Fourier and CWT spectra indicate the frequency bands of ~ 0.00029 to 0.001 $m^{-1}$ and ~ 0.00045 to 0.0001 $m^{-1}$, respectively. That is, the Fourier spectrum of wavelengths from ~1000 to 3333 m and CWT spectrum wavelengths from ~ 1000 to 2222 m. The FWHM is wider in standard Fourier spectrum because of the intersection curves of the power spectrum of high land slope (potential landslide areas) and stable (plain areas). Extracting on Google Earth$^{TM}$ the maximum altitude in the region it was found that the highest value is 2105 m. Thus, the 2D-CWT got improved precision (standard deviation of 5.5%).

## 4. Mapping Risk Areas

In order to map terrains with characteristics indicative of risk areas, the spatial patterns of spectral power contained in the frequency bands were analyzed with the data in the last section and using the transforms of Section 2. As wavelet analysis preserves the information about scale and position, Eq. (1) was directly implemented to map spatial patterns of spectral power with 2D-CWT. The algorithm first computes a wavelet coefficient at each point of DEM for each specified scale wavelet. Then, the quadratic sum of wavelet coefficients is computed at each point and the result is displayed. A point with a high value indicates a characteristic topographical landslide deep that area.

The arrays produced by the application of algorithms 2D-CWT and 2D-DFT for the areas analyzed clearly highlight variations in the spectral power contained in the wavelength characteristic slope across the ground area (Fig. 6). The sum of the spectral power varies in magnitude in powers of 10 in each studied area, reflecting the considerable variation in the topography of the terrain. The analyzed areas that are smooth across the range of spatial frequencies (Fig. 5) have low spectral power, while areas that are steep have high spectral power over this frequency band. These areas of high spectral power tend to coincide with the areas detected as having high risk of landslides, corroborating the assumption that the total spectral power can delineate terrain landslide. Due to the improvement in the wavelet transform (energy unit in time and spectrum), peaks become less dispersed even when applied to variance in the CWT transformed (in order to avoid the edge effect). Previous results often entailed the fusion of hills, hiding the crucial information of potential landslide areas.

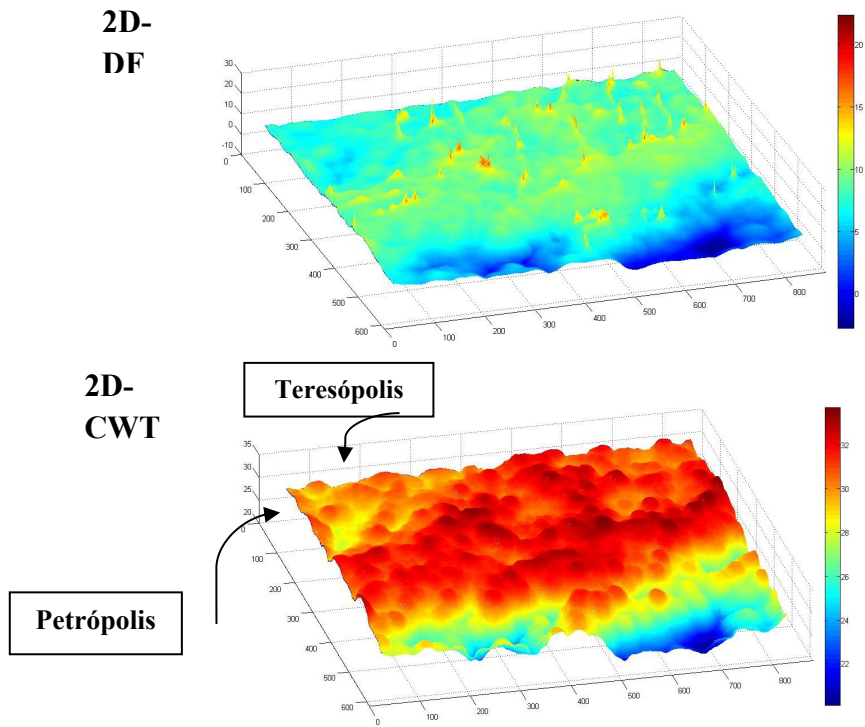

**Fig. 8**. Power Spectral sums of (top) 2D-DFT and (bottom) 2D-CWT of the study area.
Colors above the red (dark wine) indicate signs of intense land topography with deep slopes.
Colors toward blue indicate more stable and flat terrain or when it is normally intense blue of the sea or rivers.

The comparison between the natural terrain of the region and the detection of areas with deep slopes from slipping, the 2D-CWT showed better signs from the slopes of the escarpment. The 2D-DFT spotted only small dots in areas of steep slope, but it has the advantage of making the result less dispersed. That was not the case here, but tests [6] in areas where the terrain with high slope approaches the ground stable results were similar between the two algorithms. This is due to the fact that the analytical range of wavelengths in this situation became narrower.

The topographical features of the studied areas with sharp features tend to be wrongly classified as slopes with high risk of land sliding, as for example the mountain region in the area north of the State of Rio. Those spikes produce high energy spectral peaks in the Fourier analysis are necessary because large amplitudes of sinus and cosines to capture sudden changes in the analyzed data. In other words, any analyzing window that contains an abrupt change in altitude is therefore a sum of spectral energy abnormally high. The analysis with Mexican hat or Airy wavelets also tends to highlight edges because of their high inclination.

## 5. Conclusions and Lines for a Further Work

Maps with inventory risk areas provide valuable information to scientists and policy-making, but its creation remains a time-consuming and subjective. The growing availability of high-resolution DEMs derived from LiDAR offers new opportunities to improve efficiency and objectivity of mapping spatial and temporal patterns of unstable slopes. However, few studies until now have explored LiDAR derived DEM to investigate landslides in resolutions on the scale of meters rather than stripes of kilometers of land.

Both techniques of digital image processing, shading, and 2D-CWT can be used to quantify morphological characteristics using high-resolution topographic data, and map how these characteristics vary spatially. Power spectra produced by 2D-Fourier and 2D-CWT analysis define the characteristic wavelengths of landslide areas within the areas of study. The sum of the power spectrum over a particular band of wavelengths provides a measure of the intensity of the characteristic topographical with risks of slippage at any point in the dataset in hundreds of square kilometers.

The 2D-CWT applied on the image of areas with big altitude difference between the flat land and areas of steep slopes yielded greater precision in the "altimetry mapping" of the region.

The approach developed here is wide-ranging and can be used to guide the search of land slide risk areas depending on surface contours. Nevertheless, further information, such as the kind of soil and vegetation, existing constructions on it, etc., is also needed to better evaluate the risks involved. In the case of the State of Rio de Janeiro (Brazil) there is already some information that can be used to better evaluate the accident risks involved. Unfortunately, such information is still not incorporated in DEM maps yet.